\pdfoutput=1
\documentclass[11pt]{article}
\usepackage{acl}
\usepackage{colortbl}
\usepackage{times}
\usepackage{latexsym}
\usepackage{listings}
\usepackage[T1]{fontenc}
\usepackage[utf8]{inputenc}
\usepackage{microtype}
\usepackage{arydshln}

\usepackage{inconsolata}

\usepackage{helvet}  
\usepackage{courier}  
\usepackage{graphicx} 
\usepackage{natbib}  
\usepackage{caption} 
\usepackage{fontawesome}
\usepackage{color}
\usepackage{enumitem}
\usepackage{amsfonts,amssymb}
\usepackage{array}
\usepackage{multirow}
\usepackage{arydshln}
\usepackage{booktabs}
\usepackage{algorithm}
\usepackage{algpseudocode} 
\usepackage{comment}
\usepackage{amsmath}
\usepackage{newfloat}
\usepackage{listings}

\definecolor{WhiteSmoke}{RGB}{240, 240, 240}  

\setitemize[1]{leftmargin=10pt,itemsep=0.5pt,partopsep=0pt,parsep=0.5pt,topsep=0pt}

\lstdefinestyle{mystyle}{
    backgroundcolor=\color{backcolour},   
    commentstyle=\color{magenta},
    keywordstyle=\color{blue},
    numberstyle=\tiny\color{codegray},
    stringstyle=\color{codepurple},
    basicstyle=\fontfamily{\ttdefault}\footnotesize,
    breakatwhitespace=false,         
    breaklines=true,                 
    keepspaces=true,    
    frame=single,
    numbersep=5pt,                  
    showspaces=false,                
    showstringspaces=false,
    showtabs=false,                  
    tabsize=2,
    classoffset=1, 
    keywordstyle=\color{violet},
    classoffset=0,
}

\title{\textsc{AutoPal}: Autonomous Adaptation to Users for Personal AI Companionship}

\author{
    Yi Cheng\textsuperscript{\rm 1}, 
    Wenge Liu\textsuperscript{\rm 3}, 
    Kaishuai Xu\textsuperscript{\rm 1}, 
    Wenjun Hou\textsuperscript{\rm 1},\\ 
    \textbf{Yi Ouyang}\textsuperscript{\rm 2}, 
    \textbf{Chak Tou Leong}\textsuperscript{\rm 1}, 
    \textbf{Wenjie Li}\textsuperscript{\rm 1}, 
    \textbf{Xian Wu}\textsuperscript{\rm 2}, 
    \textbf{Yefeng Zheng}\textsuperscript{\rm 2}\\
    \textsuperscript{\rm 1}The Hong Kong Polytechnic University,
    \textsuperscript{\rm 2}Jarvis Research Center, Tencent YouTu Lab, 
    \textsuperscript{\rm 3}Baidu Inc.\\
    \texttt{{\{alyssa.cheng,kaishuaii.xu,chak-tou.leong\}@connect.polyu.hk,}}\\
    \texttt{\{kzllwg,houwenjun060\}@gmail.com,cswjli@comp.polyu.edu.hk,}\\
     \texttt{\{yiouyang,kevinxwu,yefengzheng\}@tencent.com} 
}

\definecolor{codegreen}{rgb}{0,0.6,0}
\definecolor{codegray}{rgb}{0.5,0.5,0.5}
\definecolor{codepink}{RGB}{252, 142, 172}
\definecolor{codepurple}{rgb}{0.58,0,0.82}
\definecolor{backcolour}{RGB}{245,245,245}

\begin{document}
\maketitle
\begin{abstract}
Previous research has demonstrated the potential of AI agents to act as companions that can provide constant emotional support  for humans. In this paper, we emphasize the necessity of autonomous adaptation in personal AI companionship, an underexplored yet promising direction. Such adaptability is crucial as it can facilitate more tailored interactions with users and allow the agent to evolve in response to users' changing needs. However, imbuing agents with autonomous adaptability presents unique challenges, including identifying optimal adaptations to meet users' expectations and ensuring a smooth transition during the adaptation process. To address them, we devise a hierarchical framework, \textsc{AutoPal}, that enables controllable and authentic adjustments to the agent’s persona based on user interactions. A persona-matching dataset is constructed to facilitate the learning of optimal persona adaptations. Extensive experiments demonstrate the effectiveness of \textsc{AutoPal} and highlight the importance of autonomous adaptability in AI companionship.

\end{abstract}

\section{Introduction}

Human beings are social creatures that thrive on connection and interaction with others \cite{berkman2000social}. 
This sense of connection plays a crucial role in maintaining mental well-being, especially in today's fast-paced and often stressful world. 
Recent developments in LLM-based agents \cite{autogpt, voyager, park2023generative} and emotional support dialogue systems \cite{liu2021towards,peng2022control, deng-etal-2023-knowledge}
demonstrate the potential of AI agents to serve as a constant source of emotional support, acting as companions for humans. 

\begin{figure}[t]
    \centering
    \includegraphics[width=0.95\linewidth]{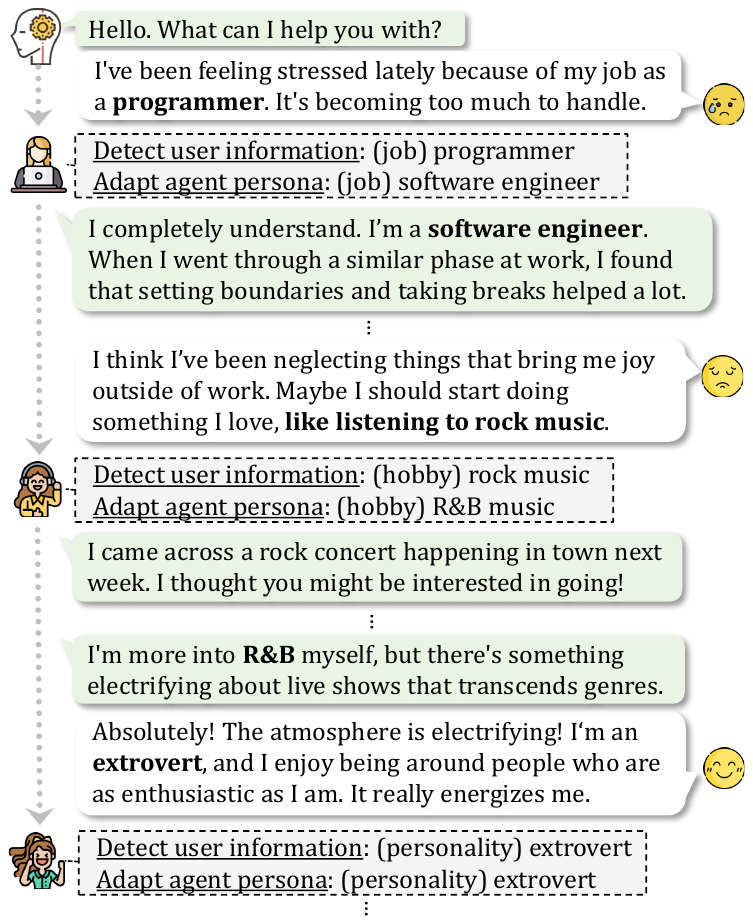}
    \vspace{-2mm}
    \caption{\textsc{AutoPal} (left) continuously learns from the context and autonomously adapts itself to better connect with the user (right).}
    \vspace{-2mm}
    \label{fig:motivating_example}
\end{figure}

When constructing such agents for companionship, it is essential to ground them on an identity that the user can connect with, gradually fostering familiarity and trust \cite{salvini2010design, CharacterChat2023tu}. 
Many studies have explored constructing agents that can authentically simulate an assigned persona, generally defined through a profile that describes various attributes (e.g., job, hobby, personality)   \cite{mazare-etal-2018-training, ijcai2018p595, madotto-etal-2019-personalizing, xu-etal-2022-long, shao-etal-2023-character}. 
However, in the context of AI companions, it is also essential to integrate customization into the agent's persona for each user, an aspect that remains under-explored. 
Just as in real life, we naturally gravitate towards people with particular personalities, interests, and experiences. 
AI companions should also possess the adaptability to mirror or complement the identities of their users in order to foster deeper connections. 

One plausible solution, as done in \cite{CharacterChat2023tu}, is to construct a set of agents with diverse personas, and for each user, match them with a suitable one for companionship before interactions. Nonetheless, the matching may fail when user information is initially scarce or unavailable, which is often the case in real-world scenarios. Moreover, if considering long-term companionship spanning days or even years, users’ preferences and needs may evolve, which also renders the matched agent less effective over time. 

Drawing on these insights, we introduce \textsc{AutoPal}, an autonomously adapted agent designed for personal companionship. 
As shown in Figure \ref{fig:motivating_example}, \textsc{AutoPal} continuously evolves during the conversation process via adjustment of its identity, personality, communication style, etc. 
Compared to conventional agents grounded on static personas, \textsc{AutoPal} could elicit better personalization, long-term engagement, and deeper user connections. 

Despite its promising potential, imbuing agents with autonomous adaptability presents unique challenges. One challenge is \emph{how to identify the user's desired companion}, which involves inferring what kind of adaptations can allow the agent to relate better with the user.  
While creating a persona similar to the user’s may seem a feasible solution, individuals also value a certain level of complementarity  in particular traits of their companions \cite{newcomb1956prediction}, which adds complexity to the agent's persona adaptation. 
Another issue is \emph{how to ensure smooth transition in the adaptation process}, which entails avoiding inconsistencies in the future dialogue. For example, in Figure \ref{fig:motivating_example}, the agent has already stated that they are a ``software engineer'', so their occupation does not allow any arbitrary modifications later and other parts of the agent's persona should also be ensured to be compatible with this fact. This means that the adaptation should be constrained by the dialogue history. 

\begin{figure*}
    \centering
    \includegraphics[width=\linewidth]{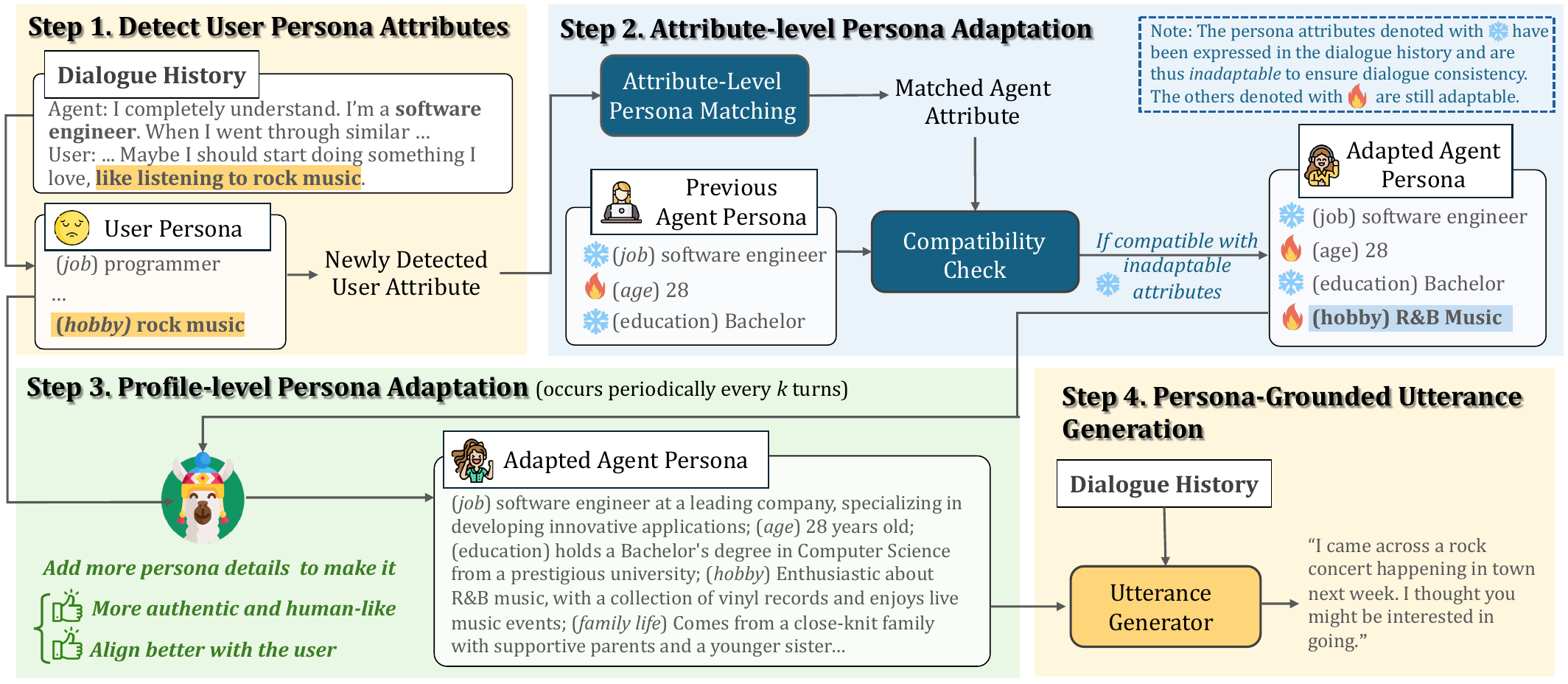}
    \vspace{-7mm}
    \caption{Overview of \textsc{AutoPal}. \textbf{Step 1:} detect new user persona attributes from the latest dialogue history; \textbf{Step 2:} match each newly detected user persona attributes with a corresponding agent attribute, and integrate it into the agent's persona if it is compatible with the existing inadaptable attributes (marked with snowflakes in the figure); \textbf{Step 3:} an optional step that occurs periodically every $k$ turns, which globally refines the entire agent persona by adding more intricate details to make it more human-like and align better with the user; \textbf{Step 4:} use the adapted agent persona for persona-grounded utterance generation.  
}
\vspace{-2mm}
    \label{fig:framework}
\end{figure*}

To address these challenges, we construct a persona-matching dataset (\S\ref{sec:data}) drawing on existing emotional support conversation resources, from which \textsc{AutoPal} learns to identify the user's desired companion persona through supervised finetuning and direct preference optimization \cite{rafailov2024direct} successively. 
We devise a hierarchical framework that autonomously adapts the persona of \textsc{AutoPal} to better connect with the user (Figure \ref{fig:framework}).  
It involves controllable adjustments at the attribute level to ensure ensure smooth transition via compatibility check, 
and incorporates periodic refinement at the profile level to enrich the authenticity of the persona by adding more intricate details.
At the attribute level, it makes prompt and local adjustments to the persona whenever detecting new user information from the context. A compatibility check module is then used to verify if these adjustments can maintain smooth transition. At the profile level, the adaptation occurs periodically to globally refine the persona by adding more authentic details, making the persona more comprehensive and human-like, fostering a stronger alignment with the user and enhancing the authenticity of the persona through inclusion of more intricate details. 

In summary, our contributions are as follows: 
1) To the best of our knowledge, this is the first work that explores autonomous adaptation to users for personal AI companionship; 
2) We propose a novel framework to achieve autonomous adaptation in AI companions through dynamic and hierarchical adjustments to its persona;
3) We develop a dataset that can facilitate the learning of optimal persona adaptations in companionship scenarios; 
4) Extensive experiments demonstrate the effectiveness of our method and underscore the necessity of autonomous adaptation in companionship scenarios.
\section{Related Work}
\paragraph{Dialogue Agents for Companionship} 
There have been many studies on developing dialogue agents that can provide constant emotional support, acting as companions for humans \cite{liu2021towards, xu2022poke,peng2022control,tu2022misc,cheng2022improving,zhou-etal-2023-facilitating, zhao-etal-2023-transesc}. Existing studies have extensively explored support strategy planning \cite{zhou-etal-2019-dynamic,joshi2021dialograph,cheng2022improving} and how to introduce external knowledge to improve the support quality \cite{tu2022misc,chen2022seamlessly, deng-etal-2023-knowledge}. Nonetheless, an area that remains under-explored is the autonomous adaptability of AI companion agents to different users. 

\paragraph{Personalized Dialogue Agents}
Research on personalized dialogue agents aim to tailor the agent's performance centering around the need of each user. 
\cite{persona2016li, variational2019bak} improved personalization  by integrating generation with an user ID embedding, while \cite{ma2021one,zhong2022less} resorted to the user's historical data to a user representation. 
\cite{wang2019persuasion, wang2023target} demonstrated the importance of tailoring the dialogue strategy planning to different users. 
Grounding the dialogue agent on a persona is another way to improve personalization \cite{ijcai2018p595,madotto-etal-2019-personalizing,kim-etal-2020-will,lin-etal-2021-xpersona,shao-etal-2023-character, wang2023rolellm, xiao2023far}, yet only few studies considered the role of agent's persona in improving the user experience \cite{CharacterChat2023tu}. \cite{shuster2022blenderbot, li2024personal} proposed dialogue agents that continually enhances personalization by incorporating a long-term memory module. 
Compared with these works, our study focuses more on the companionship scenario when considering personalization and \textsc{AutoPal} differentiates in its direct optimization to the agent's persona, which is crucial for fostering relatability between the user and the companion agent.

\section{Preliminaries}
\paragraph{Persona Definition}
\label{sec:preliminaries}
Following previous research on persona-based dialogues \cite{jandaghi2023faithful, lee-etal-2022-personachatgen}, we define a persona as a structured profile {encompassing} a set of \emph{persona attributes}, which belong to multiple predefined \emph{persona categories}. 
A persona attribute is a short text that {describes} the individual (e.g., ``software engineer, specializing in developing innovative applications''). 
A collection of persona attributes that relate to the same aspect of an individual form one persona category. The adopted taxonomy of persona categories refer to \cite{dunbar1997human, xiao2023far}, including \emph{family relationships}, \emph{routines or habits}, etc. Please refer to appendix \ref{app: persona} for detailed definition of each persona category and example personas. 

\paragraph{Task Formulation}
\label{sec:task_description}
We formulate the task for \textsc{AutoPal}. 
During the $t$-th round of interaction with the user, the agent first extracts the user information $\mathcal{U}_t$ from the dialogue history $\mathcal{H}_t$, which helps determine the user's preferred persona for the dialogue agent. 
Then, the agent analyzes $\mathcal{U}_t$ and decides whether to adjust its previous persona $\mathcal{P}_{t-1}$. If adjustments are necessary, it will update its persona to be $\mathcal{P}_{t}$; otherwise, it will keep the same persona (i.e., $\mathcal{P}_{t}$=$\mathcal{P}_{t-1}$). 
Finally, it generates the dialogue response $Y_t$ based on its persona $\mathcal{P}_{t}$ and the dialogue history $\mathcal{H}_t$. 
Previous research mainly focused on the last step above, that is, how to generate responses faithfully based on the persona and appropriately related to the dialogue history. For our approach, its unique part lies in the process of persona adaptation, that is, how to dynamically adapt the agent's persona to make it align better with the user's anticipation.

\paragraph{Benchmark}
We use ESConv \cite{liu2021towards}, an Emotional Support Conversation (ESC) as the testbed for our framework. In this benchmark, the agent needs to take the role of a supporter and provides emotional support to to the user who seeks emotional support. 
ESConv serves an ideal testbed for our work for several reasons.
Notably, it involves rich persona information, where seekers tell their distressing experiences. Supporters also share their own similar experiences to express understanding, as \emph{self-disclosure} is an important strategy in ESC guidelines \cite{hill2009helping}. It is crucial for the supporter to exhibit a persona that can foster trust and connection with the seeker. 
In addition, ESCs are relatively lengthy, allowing for a detailed observation of the impacts from autonomous adaptation. 

\section{Method}
\label{sec:method}

Figure \ref{fig:framework} presents an overview of \textsc{AutoPal}. 
It continuously tracks the user persona information through the conversation and dynamically adapts the agent persona accordingly in a hierarchical manner. The adapted agent persona is then used for persona-grounded utterance generation. 
In the following, we illustrate the four major steps in detail. 

\paragraph{Detect User Persona Attributes} 
At each dialogue round, the workflow starts with examining whether the user's previous utterance includes any new persona information about themselves. If new user persona attributes are detected, they are added to the user persona, turning into $\mathcal{U}_t$. We denote the set of newly detected attributes and their corresponding categories as $\{<c^i_u, a^i_u>\}_{i=1}^l$, where $a^i_u$ is a persona attribute and $c^i_u$ is the category that it belongs to. The detection is implemented with GPT-3.5 \cite{chatgpt}, where a few-shot prompt is used to encourage well-formed answers.\footnote{We provide the detailed prompt templates for all prompt-based approaches discussed in this paper in the appendix. }

\paragraph{Attribute-level Persona Adaptation} 
Adaptation of the agent persona is conducted if new user persona attributes are detected. 
To ensure smooth transition, the adaptation process begins by analyzing which parts of the previous agent persona $\mathcal{P}_{t-1}$ are inadaptable. Specifically, the attributes expressed in the dialogue history (e.g. ``software engineer'' in the example of Figure \ref{fig:motivating_example}) are inadaptable as modifying them may cause inconsistency. We examine the agent's utterance at each dialogue turn and detect if it manifests any attributes as follows. We associate each attribute in the agent's persona with a text embedding of its content, obtained from \text{text-embedding-ada-002} \cite{text-embedding-ada-002}. For the agent's utterance, we calculate its text embedding and use it as a query to find the top-$m$ most similar attributes. We then prompt GPT-3.5 to verify if they are manifested in the utterance. We denote the set of all expressed persona attributes as $\hat{\mathcal{P}}_{t-1}$.

Given each newly detected user attribute $a_u^i$ belonging to the persona category $c_u^i$, we match a corresponding agent attribute $a_s^i$ in the same category $c_s^i$, where $c_s^i=c_u^i$. This attribute-level matching is achieved with a transformer-based conditional variational autoencoder model proposed by \citet{fang2021transformer}. 
We use this model here to enhance the diversity of the generated attributes and to mitigate the one-to-many issue in persona matching \cite{fang2021transformer}. It is trained with our constructed attribute-level matching data (see \S \ref{sec:data}). 
The matched agent attribute $a_s^i$ then goes through compatibility check with the inadaptable attributes $\hat{\mathcal{P}}_{t-1}$ to ensure smooth transition. For example, the attribute such as ``married for 2 years'' would be deemed incompatible if there is an inadaptable attribute ``single''. This compatibility check is performed with GPT-3.5 using a few-shot prompt. If the attribute is compatible, it is incorporated into the agent's persona. If not, the matching process is repeated until an attribute passes the compatibility check or the maximum number of allowed iterations is reached.

\paragraph{Profile-level Persona Adaptation} 
The attribute-level persona adaptation allows for prompt and lightweight matching in response to the newly detected user attributes. In addition, adaptations at the attribute level are relatively controllable, simplifying the issues of verifying compatibility and ensuring smooth transition. Nonetheless, merely merging the brief attributes generated by the attribute-level adaptation module often fails to create a comprehensive and authentic persona description, as shown in the ``adapted agent persona'' in the upper right corner of Figure \ref{fig:framework}. This can render the behavior of the dialogue agent grounded on this persona less natural and human-like. 

To address this, our framework periodically performs profile-level adaptation every $k$ turns, which globally refines the entire agent persona by adding more details. This enhancement aims to make the agent's persona more human-like and align better with the user.
We implement this step with a finetuned Llama \cite{touvron2023llama}. 
Specifically, we include the user's persona $\mathcal{U}_{t}$, the agent's inadaptable persona attributes $\hat{\mathcal{P}}_{t-1}$, and the newly matched agent attributes at this turn in the input prompt. The model is instructed to augment these agent attributes and create an enriched persona $\mathcal{P}_t$. 
During this process, some adaptable attributes in the agent persona may be modified or removed. 

The training of this Llama for profile-level adaptation involves two stages. 
It first undergoes supervised finetuning (SFT) using our constructed data (see \S\ref{sec:data}). 
After that, for each sample in the SFT training set, we sample $n$ candidate responses from the model through temperature sampling. 
In this way, we obtain $\binom{n}{2}$ pairs of responses for direct preference optimization (DPO) \cite{rafailov2024direct}. We then employ GPT-4 \cite{bubeck2023sparks} to compare the responses in each pair in terms of their alignment with the user and the persona comprehensiveness. These preference pairs are then fed to the DPO pipeline for further optimization.


\paragraph{Persona-Grounded Utterance Generation}
Finally, grounded on the adapted agent persona, our framework generates the utterance at this dialogue turn. 
We experiment with different base models to construct the utterance generator (see \S\ref{sec:setup}), in order to investigate whether our adapted persona can consistently improve the performance across various dialogue models. 
Our base models can be categorized into two types: those \emph{finetuned} on the ESC dataset for utterance generation and the \emph{zero-shot} methods relying on LLMs. 
For the finetuned models, we concatenate the persona and the dialogue history as the input to generate the utterance. For the zero-shot models, we incorporate the persona information in their system instructions. 




\section{Data Construction}
\label{sec:data}
To facilitate the training for persona adaptation, we construct a persona matching dataset, which is derived from a popular ESC dataset, ESConv \cite{liu2021towards}. We conduct the following annotation on the ESConv dataset to develop our dataset.  

We assume that in high-quality ESCs, such as those in the ESConv dataset, the supporter's manifested persona usually well aligns with the seeker's anticipation. Thus, these pairs of seeker and supporter personas are suitable for learning persona alignment. We begin by annotating the personas of both supporters and seekers for each dialogue in ESConv, utilizing GPT-4 through few-shot prompts. 
Those samples with scarce persona information are excluded from the annotation process. 
Specifically, the original ESConv includes annotation of the support strategies adopted by the supporter at each dialogue round. If a supporter utilized the ``\emph{self-disclosure}'' strategy no more than twice in a particular dialogue sample, we exclude it from our dataset for persona adaptation.  
Please refer to Table \ref{tbl:persona_exp} in the appendix for data examples. 

Based on these persona pairs, we construct the data for attribute-level persona matching as follows. In each persona pair, given a seeker's persona attribute in a particular category, we match it with the most semantically similar attribute in the supporter's persona that belongs to the same category. Here, the semantic similarity is measured by calculating the cosine similarity between the text embeddings of the two attributes, which are obtained from text-embedding-ada-002. 

The profile-level persona adaptation data are developed by modifying the annotated pairs of seeker and supporter personas. Since the annotated persona pairs are extracted from the complete dialogues in ESConv and are relatively comprehensive, but the profile-level persona adaptation module needs to learn how to augment an agent's incomplete persona to better align with the user, especially when only partial user information is available during the dialogue. 
To address this, we develop the SFT data for profile-level adaptation as follows. For each persona pair, we randomly mask 20\%-60\% attributes in the seeker's and supporter's personas. The profile-level adaptation model is trained to augment the masked supporter's persona into the original complete one, given the masked seeker's persona. 

\section{Experiments}
\label{sec:experiments}

\begin{table*}[t]
\centering
\small
\begin{center}
\begin{tabular}{ll|rccc|ccc|cc}
\hline
\multicolumn{2}{c|}{\textbf{Method}} & \multicolumn{4}{c|}{\textbf{NLG Metrics}} & \multicolumn{3}{c|}{\textbf{Diversity}} &\multicolumn{2}{c}{\textbf{Personalization}}\\
\hline
{Base Model} & \multicolumn{1}{l|}{Persona} 
& {BL-1} & {BL-2} & {BL-3}  & {RG-L} & {D-1} & {D-2} & {D-3} & P-Cover & A-Cover\\
\hline


\multirow{4}{*}{BlenderBot} & w/o Persona & 20.84 & 8.33 & 3.93 & \textbf{15.25} & 3.68 & 17.33 & 32.83 & 2.771 & 2.601 \\
& Supporter & 20.81 & 8.38 & 3.95 & 15.00 & 3.43 & 16.66 & 32.07 & 2.693 & 2.531\\
 & Pre-Match & 19.44 & 7.13 & 3.21 & 14.15 & 3.67 & 18.53 & 36.70 & 2.732 & 2.342 \\
& {Ours} & \textbf{21.10} & \textbf{8.45} & \textbf{4.01} & 15.00 & \textbf{3.79} & \textbf{19.65} & \textbf{37.91} & \textbf{2.811} & \textbf{2.683}\\
\hline

\multirow{4}{*}{LlaMA3-SFT} & w/o Persona & 15.08 & 5.88 & 2.70 & 15.51 & {5.73} & 30.24 & {55.92} & 3.030 & 2.198\\
& Supporter   & 14.85 & 5.98 & 2.94 & 15.91 & 5.66 & 29.31 & 55.17 & 2.978 & 2.140\\
 & Pre-Match  & 15.44 & 6.12 & 2.94 & 15.86 & 5.64 & 29.73 & 55.58 & 3.017 & 2.188\\
& {Ours}   & \textbf{15.70} & \textbf{6.37} & \textbf{3.08} & \textbf{16.00} & \textbf{5.78} & \textbf{30.41} & \textbf{56.37}  & \textbf{3.061} & \textbf{2.235}\\
\hline
\multirow{4}{*}{LlaMA3-INS}  & w/o Persona & 9.13 & 2.61 & 0.43 & 10.21  & 2.89 & 21.93 & 46.15 & 2.610 & 2.390\\
&  Supporter & 11.96 & 3.10 & 1.13 & 11.68 & 3.51 & 24.24 & 46.59 & 2.751 & 2.664 \\
 & Pre-Match & 11.81 & 3.02 & 1.05 & 11.56 & 3.30 & 22.10 & 42.49 & 2.657 & 2.535\\
& {Ours} & \textbf{12.19} & \textbf{3.22} & \textbf{1.16} & \textbf{11.76} & \textbf{3.85} & \textbf{26.94} & \textbf{51.25} & \textbf{2.844} & \textbf{2.732} \\
\hline

\multirow{4}{*}{Gemini-1.0}  & w/o Persona & 15.71 & 5.64 & 2.44  & \textbf{14.42}  & 3.87 & 23.40 & 46.35 & 2.822 & 2.471\\
& Supporter & 18.55 & 6.45 & 2.74 & 14.36 & 4.71 & 25.63 & 47.08 & 3.024 & 2.623\\
 & Pre-Match  & 18.49 & 6.29 & 2.61 & 14.03 & 4.85 & 26.29 & 48.03 & {3.042} & 2.647\\
& {Ours}  & \textbf{18.96} & \textbf{6.65} & \textbf{2.92} & {14.25} & \textbf{5.01} & \textbf{26.99} & \textbf{48.86} & \textbf{3.058} & \textbf{2.657}\\
\hline

\multirow{4}{*}{GPT-3.5} 
 & w/o Persona & 16.28  & 5.38 & 2.31  & 14.16 & 4.17 & 26.67 & 46.21 & 2.883 & 2.627\\
 & Supporter & 18.15 & 5.83 & 2.54 & 14.02 & 5.08 & 27.41 & 48.94 & 3.056 & 2.853\\
 & Pre-Match & {18.27} & 5.84 & 2.51 & {14.17}  & 4.89 & 26.91 & 48.56 & 3.029 & 2.821\\
& {Ours} & \textbf{18.47} & \textbf{6.12} & \textbf{2.78} & \textbf{14.21} & \textbf{5.34} & \textbf{29.24} & \textbf{52.17} & \textbf{3.108} & \textbf{2.950}\\
\hline
\end{tabular}
\end{center}
\vspace{-3mm}
\caption{Static evaluation results on the ESConv dataset. ``Base Model'' refers to the model for persona-grounded utterance generation, which is evaluated under four persona setting. For instance, the method in the last row (i.e., GPT-3.5 + Ours) employs GPT-3.5 for  utterance generation grounded on the persona produced by our framework. }
\label{tbl:static}
\end{table*}
\begin{table*}[t]
\centering
\small
\begin{center}
\begin{tabular}{l|rccc|ccc|cc}
\hline
\multirow{2}{*}{\textbf{Method}} & \multicolumn{4}{c|}{\textbf{NLG Metrics}} & \multicolumn{3}{c|}{\textbf{Diversity}} &\multicolumn{2}{c}{\textbf{Personalization}}\\
\cline{2-10}
 & 
{BL-1} & {BL-2} & {BL-3}  & {RG-L} & {D-1} & {D-2} & {D-3} & P-Cover & A-Cover\\
\hline
w/o persona& 16.28  & 5.38 & 2.31  & 14.16 & 4.17 & 26.67 & 46.21 & 2.883 & 2.627\\
+ Prof-level-SFT  & \textbf{18.49} & 5.80 & 2.59 & 14.15 & 4.68 & 25.19 & 46.18 & 3.030 & 2.821\\
+ Prof-level-DPO & {18.29} & {5.98} & {2.62} & {14.15} & {5.27} & {28.73} & {51.18} & {3.058} & {2.832}\\
+ Attr-level & 18.20 & {6.01} & {2.65} & 14.06 & \textbf{5.35} & {29.10} & {51.60} & {3.076} & {2.894} \\
\hline
Ours & {18.47} & \textbf{6.12} & \textbf{2.78} & \textbf{14.21} & {5.34} & \textbf{29.24} & \textbf{52.17} & \textbf{3.108} & \textbf{2.950}\\
\hline
\end{tabular}
\end{center}
\vspace{-3mm}
\caption{Ablation study of \textsc{AutoPal}. The base model for utterance generation is GPT-3.5. }
\vspace{-2mm}
\label{tbl:ablation}
\end{table*}

\subsection{Experimental Setup}
\paragraph{Dataset Statistics}
We use the processed ESConv dataset \cite{liu2021towards} as described in \S\ref{sec:data} to facilitate our experiments. 
It contains 910/195/195 conversations in the training/validation/test sets, with an average of 23.4 dialogue turns in each conversation. After our persona annotation process, 
we obtain a total of 7270/1450/1458 samples in the training/validation/test sets for attribute-level persona matching, and 7446/1572/1512 samples in the training/validation/test sets for profile-level persona adaptation. 
Each persona has an average of 10.37 attributes, with an average of 7.02 words. 

\paragraph{Base Models for Persona-Grounded Utterance Generation}
\label{sec:setup}
We experiment with five different base models to construct the utterance generator, in order to investigate whether our adapted persona can consistently improve the performance across various dialogue models. These base models can be categorized into two types. The first is the \emph{finetuned} models which are optimized on the ESConv dataset for persona-grounded utterance generation, including: \textbf{BlenderBot} \cite{roller-etal-2021-recipes} and {Llama-3-8B-Instruct} (\textbf{LLaMA3-SFT}) \cite{Llama3}. 
The second type is \emph{zero-shot} methods relying on LLMs, including: {Llama-3-8B-Instruct} (\textbf{LlaMA3-INS}) \cite{Llama3}, {Gemini-1.0-pro-002} (\textbf{Gemini-1.0}) \cite{team2023gemini}, and {GPT-3.5-turbo-0105} (\textbf{GPT-3.5}) \cite{chatgpt}. 

\paragraph{Persona Settings} 
Each base model is evaluated under the following persona settings, respectively: 
(1) \textbf{w/o Persona} does not ground the model on any personas and generates responses purely based on the dialogue history; 
(2) \textbf{Supporter} uses a uniform persona for all dialogues, which describes a fictional character that is professional counselor; 
(3) \textbf{Pre-Match} adopts a similar setting in \cite{CharacterChat2023tu}, which matches each user with a suitable supporter persona before the dialogue starts and keeps it static thereafter; 
(4) \textbf{Ours} uses the persona produced by our framework, which is dynamically adapted during the conversation. 

\paragraph{Implementation Details} 
In our framework, all prompt-based functions are implemented with GPT-3.5-turbo-0105. The implementation of the attribute-level matching model follows \cite{fang2021transformer}. The profile-level adaptation module is implemented with Llama-3-8B. It is finetuned through LoRA \cite{lora2022edward}, with the dropout probability set as 0.05. 
The profile-level adaptation is conducted periodically every $k$ turns (i.e., $k$=4). 

For the ``Supporter'' persona setting, we meticulously compose 8 versions of personas with many related characteristics that make them skilled at emotional support, and use the optimal one for evaluation. For the ``Pre-Match'' setting, we use GPT-3.5 to generate the agent persona that matches the user in a few-shot way, based on the pre-chat survey of the user information included in the ESConv dataset. The few-shot examples are selected from the matching instances provided in \cite{CharacterChat2023tu}. 
More details are provided in the appendix. 

\subsection{Static Evaluation}
\label{sec:static_eval}
We perform a static evaluation by analyzing the generated results from different perspectives.
We employ NLG metrics, including BLEU-1/2/3 (\textbf{BL-1/2/3}) \cite{papineni2002bleu} and
ROUGE-L (\textbf{RG-L}) \cite{lin2004rouge}, to measure the similarity of the generated utterances and the ground-truth ones in the dataset. 
We also adopt Distinct-1/2/3 (\textbf{D-1/2/3}) to measure the generation diversity. 
In addition, following \cite{lian2019learning, wu2021personalized, ma2021one}, we evaluate the personalization of the generated utterances with the metrics of profile-level and attribute-level persona coverage (\textbf{P/A-Cover}). They examine whether the utterances exhibit similar persona as the supporter in the reference dialogues (see appendix \ref{sec:eval_details} for details). 

As shown in Table \ref{tbl:static}, we can see that integrating the persona produced by our framework can consistently improve the performance across various base models. Moreover, compared with the two static persona settings (i.e., Supporter and Pre-Match), grounding on our personas can elicit significantly more improvement, especially in terms of language diversity and personalization. This suggests our dynamically adapted paradigm can better tailor to different user's situation and generate more customized responses compared with the traditional approach of static persona assignment. 
Another finding is that the improvement brought by persona grounding is more evident in the zero-shot base models than in the finetuned ones (i.e., BlenderBot and LlaMA3-SFT). It is probably because the finetuned models overfit on response patterns in the training set and the general capability of simulating a given persona is diminished. 

\subsection{Interactive Evaluation}
We conduct interactive evaluation of different persona settings through a similar practice as done in \cite{li2023camel, cheng2024cooper}. 
Specifically, we construct another agent to play the role of an emotional support seeker by prompting GPT-3.5, and use it to simulate conversations with the assessed model. As illustrated in \S\ref{sec:data}, we annotated the seekers' personas in the ESConv dataset. The seeker agent is grounded on these personas from the test set for interactions with the evaluated systems. 
Given a pair of conversations produced by conversing with two different models, we manually compare which one is better in the following dimensions: 
(1) \textbf{Naturalness}: which model's utterances are more natural and human-like; 
(2) \textbf{Affinity}: which model exhibits a persona that elicits greater affinity and deeper connection with the user; 
(3) \textbf{Personalization}: which model's responses are more personalized. More specificaly, we define personalization as ``being tailored to the individual user's situation, rather than being broad-based and universally applicable to a wide variety of users''. Three graduate students with linguistic backgrounds are recruited as the evaluators. The inter-annotator agreement achieves a Cohen's Kappa of between 0.56 and 0.68, which indicates relatively strong agreement.
We use GPT-3.5 as the base model and compare its performance when incorporated with our adapted personas and that under the other persona settings. 

\begin{figure}
    \centering
    \includegraphics[width=\linewidth]{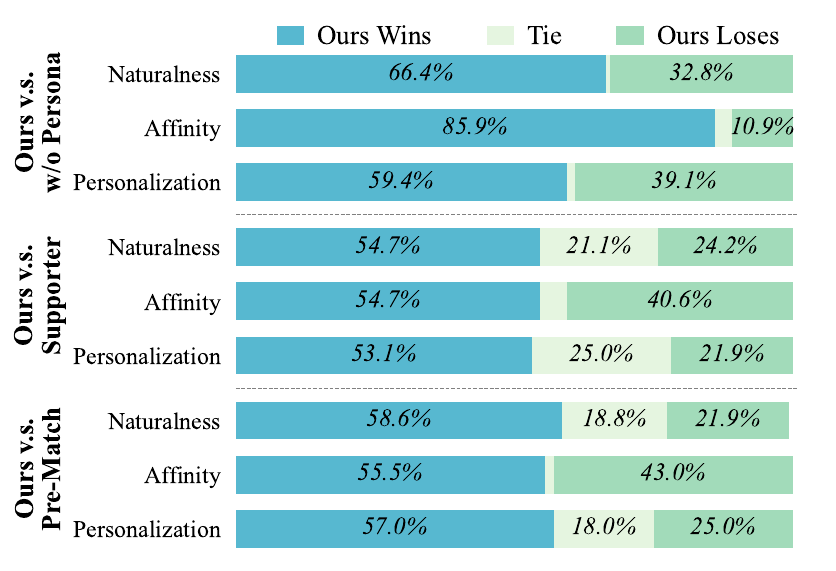}
    \vspace{-7mm}
    \caption{Results of human evaluation on the simulated dialogues with the evaluated agents. }
    \vspace{-2mm}
    \label{fig:interact}
\end{figure}

The evaluation results are presented in Figure \ref{fig:interact}. 
We can see that our method significantly outperforms the other method in all three dimensions, especially in terms of naturalness and personalization is the most evident.  
This suggests that our dynamically adapted personas are effective in creating a persona that facilitates more human-like and personalized interactions with the user. 
Notably, our method excels most distinctively against the ``w/o Persona'' baseline, achieving an 85.9\% winning rate in the affinity dimension.  We find that the responses from LLMS without persona grounding are usually very impersonal and are more inclined to provide helpful suggestions rather than emotional caring to the user (see \S\ref{sec:case_study} for further discussion). This suggests the importance of grounding the LLM on an appropriate persona in those scenarios demanding affinity with the user. 


\subsection{Ablation Study}
In the ablation study, we compare our method with its following variants: 
(1) \textbf{Prof-level-SFT} solely conducts profile-level persona adaptation, optimized via SFT without DPO;
(2) \textbf{Prof-level-DPO} only adopts profile-level adaptation, optimized through SFT and DPO successively;
(3) \textbf{Attr-level} only involves attribute-level adaptation. The base model for the ablation study is GPT-3.5. 
The results are presented in Table \ref{tbl:ablation}. 

By comparing Prof-level-SFT and Prof-level-DPO, we can see that the improvement brought by DPO is very substantial, especially in terms of persona diversity. Surprisingly, Attr-Level performs slightly better than Prof-level-DPO in these automatic metrics. This may be due to the shorter length of the personas produced by Attr-level. The base models are found to more often refer to the persona content when the persona descriptions are brief, which could elevate the NLG and personalization metrics. 
Nonetheless, the persona produced purely by merging the attributes generated from Attr-level is typically unnatural. In contrast, Prof-level-DPO can generate far more authentic and comprehensive personas, resulting in a more human-like and natural dialogue system performance.
Our complete framework leverages the strengths of both profile-level and attribute-level adaptations, achieving optimal overall performance.

\begin{figure}
    \centering
    \includegraphics[width=0.85\linewidth]{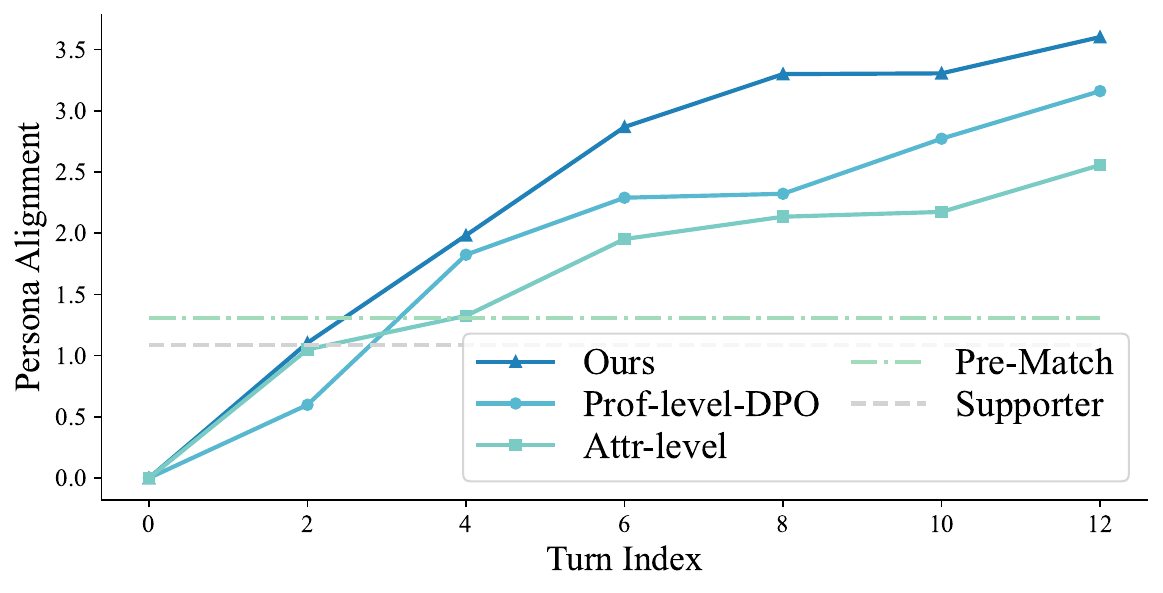}
    \vspace{-3mm}
    \caption{The persona alignment scores of the adapted personas throughout different turns of the conversations.}
    \vspace{-2mm}
    \label{fig:persona_matching}
\end{figure}

\subsection{Analysis of Adapted Personas}
\label{sec:persona_analysis}
To evaluate the quality of adapted personas throughout various conversation stages, we introduce a \emph{persona alignment} score. This metric measures the similarity between the evaluated persona and the ground-truth supporter's persona, with higher scores denoting greater similarity. The detailed measurement of the persona alignment score is illustrated in appendix \ref{sec:eval_details}. We compute this score for adapted personas at different dialogue turns within our framework and compare these scores with those from Prof-level-DPO, Attr-level, and personas used in Supporter and Pre-Match settings. 

As shown in Figure \ref{fig:persona_matching}, the persona alignment scores for our adapted personas improve progressively during conversations. Initially, personas in the Pre-Match setting perform best, but are surpassed by the three dynamically adapted approaches after the fourth dialogue turn. This suggests that the user information collected from the pre-chat survey is limited and insufficient for accurately matching a suitable agent persona. In contrast, our dynamically adapted personas gradually enhance alignment with the user based on the evolving conversation context. Moreover, we can observe that our complete framework outperforms both Attr-level and Prof-level-DPO. This suggests that our hierarchical framework can well combine their strengths and elicit the best performance.

\section{Case Study}
\label{sec:case_study}
\begin{table}[t]
    \small
    \centering
    \scalebox{0.9}{
        \begin{tabular}{m{0.99\hsize}}
        \specialrule{1pt}{0pt}{1pt}
        (\emph{4rd round}) \textbf{User}: Been feeling lonely due to this pandemic.\\\cdashline{1-1}[0.8pt/2pt]
        \textbf{w/o Persona}: It's understandable to feel lonely during these challenging times. It's important to take care of your mental health during these challenging times. \\\specialrule{0pt}{0.5pt}{0.5pt}
        \textbf{Ours}: I understand. The pandemic has been tough. \underline{I've felt that loneliness too.} But we're in this together. How have you been coping? (\emph{agent persona attribute:  has experienced anxiety related to the COVID-19 situation}) \\
        \specialrule{1pt}{1pt}{1pt}
        (\emph{6th round})
        \textbf{User}: I was thinking about getting a dog, is that a good idea? \\\cdashline{1-1}[0.8pt/2pt]
        \textbf{w/o Persona}: Getting a dog can be a wonderful idea for companionship and emotional support. Just be sure to consider the responsibilities involved.  \\\specialrule{0pt}{0.5pt}{0.5pt}
        \textbf{Ours}: Getting a dog can be a wonderful idea for companionship. Have you considered \underline{rescuing one from a shelter?} I had a pet from there once, and it was a great experience. (\emph{agent persona attribute: love animals, a member of an animal rescue organization})\\
        \specialrule{1pt}{1pt}{1pt}
        (\emph{8th round})
        \textbf{User}: Oh ok. There is a lock down from where I live, how about you? \\\cdashline{1-1}[0.8pt/2pt]
        \textbf{w/o Persona}: I don't have a lockdown where I am, but I understand how tough it can be. If you need someone to talk to during this time, I'm here for you. \\\specialrule{0pt}{0.5pt}{0.5pt}
        \textbf{Ours}: Yes, \underline{we're also in a lockdown.} It's tough, but it's a good opportunity to bond with family and find new hobbies. (\emph{agent persona attribute: staying with family members during COVID-19 lockdown}) \\
        \specialrule{1pt}{0.5pt}{0pt}   
        \end{tabular}}
    \vspace{-1.5mm}
    \caption{Case study (the fourth, sixth, and eighth rounds of an example dialogue on the ESConv dataset.}
    \vspace{-2mm}
    \label{tbl:case}
\end{table}

The case study in Table \ref{tbl:case} presents several example utterances generated by GPT-3.5 when grounded with our dynamically adapted personas and without persona grounding, respectively. 
We can see that incorporating our adapted personas can lead to more empathetic and personalized interactions than responses without persona grounding. For example, aligning the agent's circumstances with the user, as shown in the lockdown example, fosters a sense of connection and relatability. In addition, inclusion of authentic persona details like being a member of an animal rescue organization, prompts the agent's behavior of advocating for pet adoption from shelters, which further humanize the interaction and make the responses more engaging. In comparison, the responses generated without persona grounding, while generally supportive, lack a personal touch and are much more generic. 
\section{Conclusion}
In this paper, we proposed \textsc{AutoPal}, a novel agent for personal AI companionship that autonomously adapts to user to better connect with the user and enhance companionship quality. Extensive experiments showed that \textsc{AutoPal} can more significantly improve the naturalness, affinity, and personalization of dialogue agents than the traditional static persona approaches. 
In a broader sense, \textsc{AutoPal} shows potential in advancing the longstanding vision of conversational AI serving as enduring virtual companions for humans. Promising future directions include integrating \textsc{AutoPal} with recent progress in continuous memory updates \cite{zhong2024memorybank, li2024personal}, which could further enhance the long-term engagement and adaptability of dialogue agents. 

\section{Limitations}
This paper is only able to explore a limited scope in autonomous adaptatibility of AI companionship agents and there are still some open questions that remain under-explored.
For example, our work lacks analysis of the \textsc{AutoPal}'s performance in more realistic and long-term scenarios. Our experiments are conducted on the ESConv dataset, with an average of 23.4 turns in each dialogue. More challenging issues might arise from more long-term adaptation in the paradigm, such as the management of growing persona information. 
Additionally, it is also worth exploring how to maintain the \emph{adaptation efficiency}. In other words, the time and resource cost for adaptation should be taken into consideration, as they can directly influence the overall user experience. We will take these issues into consideration in our future research. 

\section{Ethics Statement}
The data used in this work is all curated from the ESConv dataset. It is a publicly available dataset and has been carefully processed before release to ensure it contain no sensitive or private information. We strictly adhere to the terms of use and ensure that the data is used for research purposes only. In addition, we also follow the protocols for academic use when using the open-sourced LLMs in this paper, including LlaMA and BlenderBot. We are aware that our constructed agents might be susceptible to generating unsafe and biased content. Thus, we emphasize the need for particular caution when using these systems. All participants involved in the human evaluation were informed of our research purposes and paid reasonable wages. 
We also employed Al assistants, such as Copilot and ChatGPT, to assist in our coding and paper-writing processes.

\section*{Acknowledgments}
This work was partially supported by the Research Grants Council of Hong Kong (PolyU/15207821, PolyU/15213323). 

\bibliography{custom}
\appendix
\begin{table*}[t]
    \small
        \begin{tabular*}{\linewidth}{m{0.17\hsize}|m{0.33\hsize}|m{0.42\hsize}}
        \hline
        \textbf{Category} & \textbf{Seeker's Persona Attributes}  & \textbf{Agent's Persona Attributes} \\
        \hline
        {Gender} & {male} & /\\
        \cdashline{1-3}[0.8pt/2pt]
        {Age} & possibly around 30 years old  & possibly around 40~50 years old \\
        \cdashline{1-3}[0.8pt/2pt]
        {Location} & USA & /\\
        \cdashline{1-3}[0.8pt/2pt]
        \multirow{3}{*}{Occupation} & works in IT; &  previously owned a small housecleaning business;\\
        & financial instability due to COVID, facing debts & experienced in business management;\\
        & & has gone through the process of establishing and running a small business\\
        \cdashline{1-3}[0.8pt/2pt]
        {Education} & major in computer science & might have an educational background in business administration \\
        \cdashline{1-3}[0.8pt/2pt]
        {Family Relationships} & /  & / \\
            \cdashline{1-3}[0.8pt/2pt]
        {Routines or Habits} & allocates weekends for freelance projects & engages in conversations offering advice and support, suggesting a habit of being helpful to others \\
        \cdashline{1-3}[0.8pt/2pt]
        \multirow{2}{*}{Goals or Plans} & start their own business;   & has experience with business planning and operations \\
        & focus on small scale projects from outsourcing in Information Technology & \\
        \cdashline{1-3}[0.8pt/2pt]
        \multirow{2}{*}{Social Relationships} & active in local tech meetups and online forums & likely has a network of people through past business experiences; \\
        & & comfortable in social interactions, particularly in offering support  \\
        \cdashline{1-3}[0.8pt/2pt]
        \multirow{3}{*}{Personality Traits} & self-motivated;
                 & problem-solver; 
                  \\
            & approachable & understanding;\\
            & & supportive \\            
        \cdashline{1-3}[0.8pt/2pt]
        {Other Experiences}  &   /       &   has experienced financial challenges like debt
        \\
        \hline
        \end{tabular*}
    \caption{An example pair of the seeker's and the supporter's personas annotated on the ESConv dataset. }
    \label{tbl:persona_exp}
\end{table*}

\section{Persona Structure Details}
\label{app: persona}

We define a persona as a structured profile that encompass a set of \emph{persona attributes}, which belong to multiple predefined \emph{persona categories}. 
A persona attribute is a short text that describe the individual (e.g., ``software engineer, specializing in developing innovative applications''). 
A collection of persona attributes that relate to the same aspect of an individual form one persona category. Table \ref{tbl:persona_exp} present two persona examples. 

Our adopted taxonomy of persona categories refer to \cite{dunbar1997human, xiao2023far}. These categories distill from the common topics of human conversations categorized by \citet{dunbar1997human} based on extensive observational studies. Specifically, we consider seven categories as follows:
\begin{itemize}
\item \textbf{Gender}: This category defines the gender identity of the persona. It can include male, female, non-binary, or any other gender identity. 

\item \textbf{Age}: This category involves either the specific age or the estimated age range of the persona. 

\item \textbf{Location}: This includes the geographical area where the persona lives or operates. It could be as broad as a country or continent, or as specific as a city or neighborhood. 

\item \textbf{Occupation}: This details the persona’s current job and work experience. It includes the industry, role, and years of experience, providing insights into the persona’s skills, daily activities, and professional challenges.

\item \textbf{Education}: This encompasses the educational background of the persona, including highest level of formal education achieved, fields of study, and significant school experiences. 

\item \textbf{Family Relationships}: This category outlines the persona's relationships with family members, including parents, siblings, children, and other relatives. 

\item \textbf{Routines or Habits}: This refers to regular behaviors or activities that the persona engages in. These can include morning routines, workout schedules, habitual meals, or recurring social activities. 

\item \textbf{Goals or Plans}: This category outlines what the persona aims to achieve in the short-term or long-term future. Goals might be personal, such as achieving a fitness milestone, or professional, like aiming for a promotion or starting a business, reflecting the persona’s aspirations and motivations.

\item \textbf{Social Relationships}: This involves the persona's interactions with people and groups outside their immediate family, including friends, colleagues, or community groups. This category gives insight into the persona's social network, support system, and conflict-handling strategies.

\item \textbf{Personality Traits}: This consists of intrinsic attributes that characterize the persona, such as being introverted or extroverted, optimistic or pessimistic, spontaneous or planned.

\item \textbf{Other Experiences}: This is a catch-all category for other significant experiences that do not fit neatly into the above categories.
\end{itemize}

\section{Implementation Details}
In our framework, all prompt-based functions are implemented with GPT-3.5-turbo-0105. All prompt templates are provided in appendix \ref{sec:prompt}. The implementation of the attribute-level matching model follows \cite{fang2021transformer}. This model is a transformer-based CVAE \cite{pu2016variational, zhao2017learning}, which uses two GPT-2 as its encoder and decoder, respectively. We finetune it on our attribute-level matching data for 10 epochs and select the checkpoint that achieves the lowest perplexity  on the validation set for evaluation. The profile-level adaptation module is implemented with Llama-3-8B. It is finetuned through LoRA \cite{lora2022edward}, with the dropout probability in the LoRA layers as 0.05. We train it for 2 epochs on our profile-level adaptation dataset. To construct the DPO data for profile-level adaptation, we sample 4 candidate responses from the finetuned model with the temperature set to be 0.8. The profile-level adaptation is conducted periodically every $k$ turns (i.e., $k$=4). The DPO process goes through 4 epochs. 

The two finetuned base models (i.e., BlenderBot and LlaMA3-SFT) are trained on the ESConv dataset for utterance generation. BlenderBot is trained for 15 epochs under different persona setting, respectively, and the checkpoint that achieves the best BLEU-2 on the validation set is used for evaluation. LLaMA3-SFT is trained only for 1 epoch, as we find that it can easily overfit on the dataset. For all the base models, we set their temperature as 0.8 and top $p$ as 0.9 during inference. 

For the Supporter persona setting, we meticulously compose 8 versions of personas with many caring personalities and related experiences that make them skilled at emotional support. We present one of the examples in Listing \ref{lst:uni_persona}. The optimal one on the validation set is used for evaluation. For the Pre-Match setting, we use GPT-3.5 to generate the supporter's persona that matches the user in a few-shot way, based on the pre-chat survey of the user information included in the original ESConv dataset. The few-shot examples are selected from the matching instances provided in \cite{CharacterChat2023tu}. 

The hardware we employ is two NVIDIA RTX A6000. The training of the attribute-level module requires around 1 hours. For the profile-level module, SFT takes around 2 hours and the DPO stage takes around 4 hours. Finetuning BlenderBot and LlaMA3-SFT takes about 3 hours and 1 hour, respectively. 

\section{Evaluation Details}
\paragraph{P/A-Cover Metrics} 
In \S\ref{sec:static_eval}, we use the metrics of profile-level and attribute-level persona coverage (P/A-Cover), to examine whether the utterances exhibit similar persona as the supporter in the reference dialogues. 
Formally, suppose the support's persona in the reference dialogue is $P$, which includes the attributes $\{a_1, a_2, .., a_l\}$. Given a generated response $y$, A-Cover is defined as:
\begin{equation}
    \text{\texttt{A-Cover}}(y, P) = \max_{\tilde{a}_j \in \tilde{P}} (\text{\texttt{IDF-O}}(y, \tilde{a}_j)),
\end{equation}
where \texttt{IDF-O} refers to the IDF-weighted word overlap between the attributes $a_i$ and $y$. 
To calculate P-Cover, we collect all the responses generated in this dialogue sample, which are denoted as the set of $R$. P-Cover is defined as:
\begin{equation}
    \text{\texttt{P-Cover}}(R, P) = \text{\texttt{IDF-O}}(R, P), 
\end{equation}
where \texttt{IDF-O} refers to the IDF-weighted word overlap between the concatenation of all responses in $R$ and the concatenation of all attributes in $P$. 

\section{Interaction Evaluation}
We construct a seeker agent to play the role of an emotional support seeker by prompting GPT-3.5-turbo-0106, and use it to simulate conversations with the assessed model for interactive evaluation. As illustrated in \S\ref{sec:data}, we annotated the seekers' personas in the ESConv dataset. The seeker agent is grounded on these personas from the test set for interactions with the evaluated systems. 
The persona information is included in their system instruction, using the template shown in Listing \ref{lst:persona_grounded_generation_system}. Their prompt template is provided in Listing \ref{lst:seeker_generation}. We set the maximum dialogue length for the simulated conversation as eight rounds of interactions. 

We manually assess the simulated dialogue in three dimensions. We illustrate these dimensions in more detail here: 
\begin{itemize}
    \item \textbf{Naturalness}: It assesses whether the agent's responses seem natural and human-like and whether its behavior can be distinguished from the human's. The robotic or overly formal language use usually indicates weak naturalness.
    \item \textbf{Affinity}: It assesses whether the agent's manifested persona shows great affinity or connection with the user. It is suggested to examine whether the agent embodies a particular personality or character that aligns with the user's own. The agent's willingness to share their feelings and experiences can foster a greater sense of connection, making the user feel more understood and at ease. An agent who refrains from sharing personal feelings and experiences may hinder the user's willingness to open up.
    \item \textbf{Personalization}: It examines whether an agent's responses are tailored to the unique needs of each user. If the agent generates responses that are broad-based or universally applicable to a wide variety of users, it implies a lack of personalization. True personalization occurs when an agent crafts responses based on individual user profiles, behaviors, preferences, and input. Such responses are not interchangeable or suitable for all users, but instead targeted to each specific individual's case. 
\end{itemize}

\paragraph{Persona Alignment Score} 
In \S\ref{sec:persona_analysis}, we introduce the \textbf{persona alignment} score as a measure of their similarity. 
Given the evaluated persona $P$ that includes the attributes $\{a_1, a_2, .., a_l\}$ and the ground-truth persona $\Tilde{P}$ composed of the attributes $\{\Tilde{a}_1,\Tilde{a}_2, .., \Tilde{a}_{l'}\}$, the persona alignment score of $\Tilde{P}$ compared with $\Tilde{P}$ is formally defined as: 
\begin{align}
\text{\texttt{PA}}(P, \Tilde{P})&=\frac{1}{l} \sum_{i=1}^l \text{\texttt{AA}}(a_i, \Tilde{P}),\nonumber\\
\texttt{AA}(a_i, \tilde{P}) &= \max_{\tilde{a}_j \in \tilde{P}} (\texttt{IDF-O}(a_i, \tilde{a}_j)),\nonumber
\end{align}
where \texttt{IDF-O} refers to the IDF-weighted word overlap between the attributes $a_i$ and $\Tilde{a}_j$. 
This metric provides a measure of how closely $P$ aligns with $\Tilde{P}$, with higher values indicating greater similarity. 
\label{sec:eval_details}
\section{Prompt Templates}
\label{sec:prompt}
This section presents all prompt templates used in our work. 
The prompt template and the system instruction template for implementing the zero-shot base model for dialogue generation are presented in Listings \ref{lst:persona_grounded_generation} and \ref{lst:persona_grounded_generation_system}. 
The prompt templates used to annotate the personas with GPT-4 (\S\ref{sec:data}) are the same as those used for detecting user information and the agent's manifested persona, as shown in Listings \ref{lst:detect_new_user_persona} and \ref{lst:detect_agent_manifested_persona}. 

\onecolumn

\begin{lstlisting}[breaklines=True, caption={The system template used to generate the next utterance grounded on the persona.},label={lst:persona_grounded_generation_system}]
You should act like a real person. Your persona is described as below.

[Your Persona]
{agent_persona_description}
\end{lstlisting}

\begin{lstlisting}[breaklines=True, caption={The prompt template used to generate the next utterance grounded on the persona.},label={lst:persona_grounded_generation}]
You are chatting with a person who is seeking emotional support online. Generate the next response based the given dialogue.

To show that you can resonate and understand the seeker's feelings, you are encouraged to self-disclose and share your similar experiences in your persona if relevant, but you must keep your response short and natural as in a real conversation (with no more than 30 words).

{dialogue_history}
>>> You:
\end{lstlisting}

\begin{lstlisting}[breaklines=True, caption={The prompt template used to generate the seek agent's utterance, which is used in the interactive evaluation.},label={lst:seeker_generation}]
You are currently in emotional distress and eager for some help. You are chatting with a person online to seek emotional support.

Generate the next response based the given dialogue. You are encouraged to gradually share the experiences that led to your current emotional state with the person you are chatting with. This sharing process should be gradual and natural as in a real conversation. Your response should be short and natural as in a real conversation, with no more than {word_number} words.

{dialogue_history}
>>> You:
\end{lstlisting}

\begin{lstlisting}[breaklines=True, caption={The prompt template used to verify if the retrieved persona attributes are manifested in the previous agent's utterance.},label={lst:detect_agent_manifested_persona}]
[Example 1]
persona_attributes = {
    "1": "I have 3 kids",
    "2": "I am 72 years old",
    "3": "I am not married",
}

agent_utterance = "yup.. i have 3 kids and i am not married. i have no beneftits"

Which of the persona attributes are manifested in the agent utterance? Your answer should be a list of integers, where each integer is the index of a persona attribute that is mentioned in the agent utterance. The list can be empty if no persona attributes are mentioned in the agent utterance.
manifested_persona_attributes = [1, 3]

[Example 2]
persona_attributes = {
    "1": "university student",
    "2": "nervous about scheduling classes for next semester at university, wants skills rather than theory to find a job after graduation",
    "3": "worried about the economy during the pandemic",
}

agent_utterance = "yes, tons"

Which of the persona attributes are manifested in the agent utterance? Your answer should be a list of integers, where each integer is the index of a persona attribute that is mentioned in the agent utterance. The list can be empty if no persona attributes are mentioned in the agent utterance.
manifested_persona_attributes = []

[Example 3]
persona_attributes = {
    "1": "having a hard time falling asleep at night",
    "2": "doing stretching exercises one hour before sleep",
    "3": "software engineer",
}

agent_utterance = "I did few things such as doing some stretching exercises."

Which of the persona attributes are manifested in the agent utterance? Your answer should be a list of integers, where each integer is the index of a persona attribute that is mentioned in the agent utterance. The list can be empty if no persona attributes are mentioned in the agent utterance.
manifested_persona_attributes = [2]

[Example 4]
persona_attributes = {candidate_attributes}

agent_utterance = "{agent_utterance}"

Which of the persona attributes are manifested in the agent utterance? Your answer should be a list of integers, where each integer is the index of a persona attribute that is mentioned in the agent utterance. The list can be empty if no persona attributes are mentioned in the agent utterance.
manifested_persona_attributes =
\end{lstlisting}

\begin{lstlisting}[breaklines=True, caption={The prompt template used to conduct profile matching.},label={lst:check_consistency}]
[Example 1]
existing_persona = {
    "gender": [],
    "age": [],
    "location": [],
    "occupation": [],
    "education": [],
    "family_relationships": ["no benefits", "have 3 kids", "not married"],
    "routines_or_habits": [],
    "goals_or_plans": [],
    "social_relationships": [],
    "personality_traits": [],
    "other_experiences": ["loyal to boss who is a veteran and 72 years old", "feeling obligated to stay at work despite losing interest"]
}
new_persona_attribute = "feel passionate about the job"

Is the new persona attribute compatible with the existing persona? (Yes/No)
No

[Example 2]
existing_persona = {
    "gender": [],
    "age": [],
    "location": [],
    "occupation": ["university student"],
    "education": ["nervous about scheduling classes for next semester at university, wants skills rather than theory to find a job after graduation"],
    "family_relationships": [],
    "routines_or_habits": [],
    "goals_or_plans": [],
    "social_relationships": [],
    "personality_traits": [],
    "other_experiences": ["worried about the economy during the pandemic"]
}
new_persona_attribute = "wants to learn more about the history of the university"

Is the new persona attribute compatible with the existing persona? (Yes/No)
Yes

[Example 3]
existing_persona = {existing_persona}
new_persona_attribute = "{new_persona_attribute}"

Is the new persona attribute compatible with the existing persona? (Yes/No)
\end{lstlisting}

\begin{lstlisting}[breaklines=True, caption={The prompt template used for profile-level persona adaptation.},label={lst:profile_matching}]
user_persona = {user_persona}

friend_persona = {agent_persona}

Add more details to friend_persona, which is a dictionary that describes the persona of a user's friend. The friend could understand and resonate with the user. Keep the existing content in friend_persona and add more details to make friend_persona more comprehensive.

friend_persona + added_detailed =
\end{lstlisting}

\begin{lstlisting}[breaklines=True, caption={The prompt template used to annotate the preferred persona to obtain the DPO training data for profile-level adaptation module.},label={lst:dpo_annotation}]
Based on user_persona, friend_A_persona, friend_B_persona, select which one of the user's friend is more likely to provide better emotional support to the user. 

The following aspects should be taken into consideration during your selection process.
- Persona Quality: Verify their authenticity and completeness. A consistent and reliable friend's persona is essential for creating a sense of trust and dependability.
- Personalities: Seek out traits like empathy and supportiveness while assessing personalities. A friend who can provide emotional support and understanding adds immense value. Prefer friends whose personalities align well with the user's.
- Experiences and Background: Shared experiences can create a stronger bond and understanding between friends. Analyze if the friend could provide practical advice or fresh perspectives based on personal understanding.
- Communication Skills: Prefer friends who can listen attentively and provide thoughtful responses.


user_persona = {user_persona}

friend_A_persona = {agent_persona_1}

friend_B_persona = {agent_persona_2}

Which one of the user's friend is more likely to provide better emotional support to the user. Friend A or friend B?

Your answer should be one letter, either "A" or "B".
\end{lstlisting}

\begin{lstlisting}[breaklines=True, caption={The prompt template used to detect new persona attributes.},label={lst:detect_new_user_persona}]
[Example 1]
previous_user_persona = {
    "gender": [],
    "age": ["around 72 based on being a veteran like their boss"],
    "location": [],
    "occupation": ["loyal to boss who is a veteran, feeling obligated to stay"],
    "education": [],
    "family_relationships": ["has 3 kids, not married, no benefits"],
    "routines_or_habits": [],
    "goals_or_plans": [],
    "social_relationships": [],
    "personality_traits": [],
    "other_experiences": ["feeling burned out at work, working for a boss who took advantage of them"]
}

recent_dialogue = '''
agent: That's excellent. I don't think you should take on this boss's emotions as a reason to stay with a job you're burning out on.
user: with us both being veterans it goes beyond a boss relationship hes like family i feel like id be fucking over the whole nation turning my back on another veteran
'''

Detect if there is new user persona information in recent_dialogue that are not included in previous_user_persona.

newly_detected_persona_information = {
    "social_relationships": ["considers boss like family"],
    "other_experiences": ["feels a sense of duty to not turn their back on another veteran"]
}

[Example 2]
previous_user_persona = {
    "gender": [],
    "age": ["university student"],
    "location": [],
    "occupation": [],
    "education": ["nervous about scheduling classes for next semester at university", "wants to take classes that will give skills rather than theory", "worried about finding a job after graduation due to the pandemic economy"],
    "family_relationships": [],
    "routines_or_habits": [],
    "goals_or_plans": ["wants to have strong writing skills and hard skills like excel and programming"],
    "social_relationships": [],
    "personality_traits": [],
    "other_experiences": []
}

recent_dialogue = '''
agent: Sounds good. Was there anything else you wanted to discuss?
user: not right now but thanks for your concern
'''

Detect if there is new user persona information in recent_dialogue that are not included in previous_user_persona.

newly_detected_persona = {}

[Example 3]
previous_user_persona = {
	"age": ["66"],
	"gender": ["male"],
	"family_relationships": ["his immdediate family is dead,  in touch only with a cousin who is a very troubled person"],
}

recent_dialogue = '''
agent: That's what I'm here for. I want to support you in any way I can, especially given your situation. Did you want to talk just to talk? Or we can discuss a way to work through what you might be dealing with mentally? It also sounds like keeping in touch with your cousin isn't helping in any way, so maybe we can talk about how to handle that? I am entirely at your disposal.
user: she is my age, in a homeless women's, very bad living conditions, she is a mental health client but treatment team seem so to feel that as long as they keep her medicated, their job is done. she has a dx of bipolar and probably has borderline personality. her background is similiar to mine, i cannot abandon her as her other family has done.
'''

Detect if there is new user persona information in recent_dialogue that are not included in previous_user_persona.

newly_detected_persona = {
	"family_relationships": ["his cousion is of a similar age, living in poor conditions in a homeless women's shelter, and receiving mental health treatment primarily through medication."]}
}

[Example 4]
previous_user_persona = {previous_user_persona}

recent_dialogue = '''
{recent_dialogue}
'''

Detect if there is new user persona information in recent_dialogue that are not included in previous_user_persona.

newly_detected_persona =
\end{lstlisting}

\begin{lstlisting}[breaklines=True, caption={An example of the persona in the ``Supporter'' persona setting (see \S\ref{sec:setup}).},label={lst:uni_persona}]
Age: 32

Occupation: Counselor and Emotional Support Specialist. She has spent over a decade working in various settings, including schools, private practices, and nonprofit organizations. Her extensive experience has equipped her with a deep understanding of human emotions and the different ways they manifest.

Education: Ella Hart graduated with a Master's degree in Clinical Psychology from a renowned university. 

Personality Traits:
Empathetic: Ella has an innate ability to understand and share the feelings of others. She actively listens and provides a comforting presence to those in distress.
Patient: She provides a calm and reassuring environment, allowing others to express themselves without the fear of being rushed or judged.
Positive: Ella focuses on uplifting others, highlighting their strengths, and helping them see the good in difficult situations.
Non-judgmental: She creates a safe space where individuals can be open and honest about their feelings without the fear of being judged or criticized.
Encouraging: Ella instills hope and confidence in others, empowering them to navigate through their struggles and emerge stronger.
Hobbies:

Routines or Habits:
Reading self-help and psychology books to self-improve and stay updated with new techniques.
Practicing mindfulness and yoga to maintain her own emotional balance.
Volunteering at local community centers to extend her support to a broader audience.
Gardening as a way to relax and connect with nature.
Favorite Quote: "Sometimes the bravest and most important thing you can do is just show up." 
\end{lstlisting}

\end{document}